\def\year{2021}\relax
\documentclass[letterpaper]{article} 
\usepackage{aaai21}  
\usepackage{times}  
\usepackage{helvet} 
\usepackage{courier}  
\usepackage[hyphens]{url}  
\usepackage{graphicx} 
\usepackage{natbib}  
\usepackage{caption} 
\usepackage{subcaption}
\usepackage[normalem]{ulem}
\usepackage{enumitem}
\usepackage{bbm}
\usepackage{researchpack}
\newcommand{\explearn}{\textsc{Explearn}\xspace}

\newcommand{\cgpucb}{\textsc{cgp-ucb}\xspace}
\newcommand{\st}[1]{{\sf\color{violet}[ST: #1]}}

\newcommand{\colors}{{\sf Colors}\xspace}
\newcommand{\banknote}{{\sf Banknote}\xspace}

\title{Bandits for Learning to Explain from Explanations}
\author {
    Freya Behrens,\textsuperscript{\rm 1}
    Stefano Teso,\textsuperscript{\rm 2}
    Davide Mottin\textsuperscript{\rm 3} \\
}
\affiliations {
    \textsuperscript{\rm 1} Technische Universit\"at Berlin, Germany \\
    \textsuperscript{\rm 2} University of Trento, Italy \\
    \textsuperscript{\rm 3} Aarhus University, Denmark \\
    f.behrens@campus.tu-berlin.de, stefano.teso@unitn.it, davide@cs.au.dk
}

\begin{document}
\maketitle

\begin{abstract}
    We introduce \explearn, an online algorithm that learns to jointly output predictions \emph{and} explanations for those predictions.
    \explearn leverages Gaussian Processes (GP)-based contextual bandits.
    This brings two key benefits.
    First, GPs naturally capture different kinds of explanations and enable the system designer to control how explanations generalize across the space by virtue of choosing a suitable kernel.
    Second, \explearn builds on recent results in contextual bandits which guarantee convergence with high probability.
    Our initial experiments hint at the promise of the approach.
\end{abstract}

\section{Introduction}
\label{sec:introduction}

Nowadays, a physician who needs to discriminate between sick and healthy patients might use a black-box classifier to support her decisions.  If the classifier predicts a serious diagnosis for which invasive treatment is necessary, should the physician trust its prediction?

Tools from explanatory machine learning~\cite{guidotti2018survey} help decision makers, like the physician in our example, to understand the predictions of black-box models.  In the simplest case, these so-called \emph{explainers} unpack the reasons behind individual predictions by identifying those ``explanatory features''---namely input variables like pixels and words, or higher-level concepts like interpretable image segments---that are most responsible for a given prediction~\cite{guidotti2018local}.  
Explainers, albeit useful in evaluating the model, do not directly allow the users to \emph{correct} the model's bugs.  In our example, even if the physician noticed that the classifier's decision depends on image artifacts in a lung ultrasound, she could do little to correct the reasoning of the predictor.

As a remedy to this lack of control, we propose \explearn, a principled and general approach for learning models that \emph{explain their own predictions} and that \emph{learn from supervision about their own explanations.}

\explearn combines in an original way two key ingredients.
First, the flexibility of Gaussian Processes (GP)~\cite{rasmussen2006gaussian}, a class of non-parametric kernel-based models.  Contrary to existing techniques~\cite{teso2019explanatory,schramowski2020right,lertvittayakumjorn2020find} that are restricted to a single type of explanations, our approach accommodates different kinds of explanations by virtue of kernels.
Second, \explearn frames the interaction with the user in terms of contextual bandits and leverages principled algorithms for GP-based bandits~\cite{krause2011contextual} that guarantee convergence with high probability to a model that outputs high-quality predictions \emph{and} explanations, as long as such a model exists.

An important aspect of \explearn is that
it learns models that are self-explainable~\cite{alvarez2018towards} since a prediction is always accompanied by a corresponding explanation, without any extra steps. This is in sharp contrast with standard approaches, such as LIME~\cite{ribeiro2016should}, that extract approximate explanations using model distillation~\cite{bucilua2006model}.

In summary, we:
\begin{itemize}

    \item Introduce \explearn, an approach that combines Gaussian Processes and contextual bandits for learning self-explainable models by interacting with a domain expert.
    
    \item Discuss the conditions under which our learning task is well-founded, and in particular what kernel functions are most appropriate.
    
    \item Present initial evidence that \explearn learns good self-explainable models on a real-world and on a non-trivial synthetic dataset.

\end{itemize}

\section{Problem Statement and Conceptualization}
\label{sec:preliminaries}

We tackle \emph{learning to explain} from ground-truth explanations in an interactive setting.  

\begin{example}
Consider a physician who diagnoses sick or healthy patients and relies on a classifier to support the decisions. The classifier analyzes data about each patient and returns a prediction and an explanation.  With such information, the doctor can decide whether to trust the prediction.
\end{example}

\begin{example}
A fake news detector is a model that predicts the truthfulness of an article and at the same time why the article has a certain rating.  Based on this information, a journalist can evaluate the reliability of the prediction.
\end{example}

\noindent
In such critical scenarios, we require that the reasons behind the machine's predictions are always available to the user, who, in turn, should be allowed to correct the model whenever the predictions or explanations are wrong.

Since explanations and labels are neither readily available nor easy to access, we cast these scenarios in terms of sequential learning.  At each iteration $t = 1, 2, \ldots, T$, the machine receives an instance $x_t \in \calX := \bbR^n$ to be classified (the patient status) and returns both a \emph{prediction} $\hat{y}_t \in \calY$ (a diagnosis) and an \emph{explanation} $\hat{z}_t \in \calZ$ for the prediction (symptoms that support the diagnosis).  A human expert then provides feedback on the prediction and on how well the explanation does support the prediction.  Based on this feedback, the machine aims to learn a model that returns accurate predictions and explanations on future instances.

\subsection{A Contextual Bandits Formulation}

Our first contribution is a formulation of this problem in terms of \emph{contextual bandits}, a principled theory  for learning and optimization~\cite{cesa2006prediction,bubeck2012regret}, in which a learner observes a context, pulls an arm based on the context, and receives a reward. 
In our case, instances $x_t$ are contexts and prediction-explanation pairs $(\hat{z}_t, \hat{y}_t)$ are arms.  Upon playing an arm, corresponding to making a prediction, the machine receives a reward $f_t := f(x_t, \hat{z}_t, \hat{y}_t) + \epsilon_t$ from the user, where $\epsilon_t$ is a noise term.  The reward is proportional to the quality of the prediction and explanation.  Learning then amounts to estimating $f$ from the observed rewards.  The main challenge is how to balance \emph{exploitation} of high-quality arms and \emph{exploration} of arms with high information for learning.

Contextual bandits measure the performance with \emph{contextual regret} (or simply regret), which represents the reward lost by playing the predicted arm $(\hat{z}_t, \hat{y}_t)$ instead of the best possible arm. Regret is defined as: 
\[
    \textstyle
    r_t = \sup_{z, y} f(x_t, z, y) - f(x_t, \hat{z}_t, \hat{y}_t)
    \label{eq:reg}
\]
The cumulative regret $R_T$ of the machine in $T$ rounds is:
\[
    \textstyle
    R_T = \sum_{t = 1}^T r_t
    \label{eq:cumreg}
\]
A major advantage of this formulation is that it enables us to choose from several algorithms for contextual bandits that \emph{provably} minimize the cumulative regret and, in doing so, quickly learn how to choose high-reward arms.  Such guarantees are critical in our scenario, where explanations and labels are scarce and user interaction expensive.

\subsection{Challenges}

Effective strategies for learning to explain have to account for a number of challenges, which we briefly outline:
\begin{enumerate}
    \item \textbf{Control.} With no supervision on the explanations, the machine might produce nonsensical or biased explanations~\cite{lapuschkin2019unmasking,ross2017right,schramowski2020right}. Hence, supervision and ground-truth explanations are necessary. 

    \item \textbf{Flexibility.} No single definition of explanation exists~\cite{miller2018explanation,guidotti2018survey}.
    Different forms of explanations have been identified, e.g., relevant features~\cite{ribeiro2016should}, influential training examples~\cite{koh2017understanding}, counterfactual configurations~\cite{guidotti2018local}, and combinations of multiple approaches~\cite{bogaerts2020step}.  A proper strategy for learning to explain should accommodate different forms of explanations based on the application and  the inclinations of the user. We discuss below a list of candidate explanations for our approach.

    \item \textbf{Consistency.}  There is a tight coupling between predictions and explanations.  For this reason, the instance $x_t$, prediction $\hat{y}_t$, and explanation $\hat{z}_t$ should be semantically consistent with each other.  For instance, if $\hat{z}_t$ states that the $i$th feature of $x_t$ is irrelevant, the reward for $(x_t, \hat{z}_t, \hat{y}_t)$ should not change when the $i$th element of $x_t$ changes. As an immediate result,  $\hat{z}_t$ and $\hat{y}_t$ \emph{must} be predicted jointly.  The complex relationship between $x_t$, $\hat{z}_t$ and $\hat{y}_t$,  makes learning to explain a structured prediction problem.

    \item \textbf{Convergence.}  The most important property of a learning algorithm is that, given enough examples, it outputs a model guaranteed to produce high-quality predictions.  In our setting, this applies to the explanations produced by the model.
    
\end{enumerate}
Solving all of the above challenges is far from trivial and beyond the reach of existing approaches, which focus on relevant features and do not guarantee convergence~\cite{schramowski2020right,lertvittayakumjorn2020find}.

\section{Method}\label{sec:solution}

We can now introduce \explearn, a framework for learning self-explainable models based on contextual bandits that tackles the above challenges.
In the following we focus on binary classification $\calY {=} \{0, 1\}$,  although extensions to multi-label settings are immediate.  We also concentrate on \emph{local} explanations, which target single instances, as opposed to global (model-wide) explanations~\cite{bastani2017interpreting}.

\subsection{The \explearn Algorithm}

To minimize the regret for the contextual bandit, we devise \explearn, an algorithm that uses a principled learning strategy based on Gaussian Processes (GPs), a family of non-parametric models~\cite{rasmussen2006gaussian}.

A GP is characterized by a mean function $\mu(x)$ and a covariance function (aka kernel) $k(x,x')$ that encodes prior knowledge about the second-order properties of the functions drawn from the GP\footnote{If no evidence is given, then $\mu(x)$ can be assumed without loss of generality to be $0$.  It is also customary to consider bounded variance functions, that is, $k(x,x) \le 1$ for all $x$}.  GPs support closed-form Bayesian inference:  conditioning a GP on evidence gives another GP whose mean and covariance functions can be derived analytically.

We now instantiate GPs on our setting.  Let $\xi_t = (x_t, \hat{z}_t, \hat{y}_t)$  be the instance, explanation, and prediction handled by \explearn in the $t$th iteration and $f_t = f(\xi_t) + \epsilon_t \in [0, 1]$ be the corresponding user-supplied reward, where $\epsilon_t \sim \calN(0, \sigma^2)$ are normally distributed independent noise variables.  
The \emph{posterior GP} conditioned on the observed data $\Xi_T = \{\xi_1, \ldots, \xi_T\}$ and noisy rewards $\vf_T = (f_1, \ldots, f_T)^\top$ has the following mean $\mu_T(\xi)$ and covariance $k_T(\xi, \xi')$:
\begin{align}
    \mu_T(\xi)
        & = \vk_T(\xi)^\top \Gamma_T \vf_T
        \label{eq:mut}
    \\
    k_T(\xi, \xi')
        & = k(\xi, \xi') - \vk_T(\xi)^\top \Gamma_T \vk_T(\xi')
        \label{eq:kt}
\end{align}
where we use $\vk_T(\xi) = (k(\xi_1, \xi), \ldots, k(\xi_T, \xi))^\top$, $K_T = [k(\xi, \xi') : \xi, \xi' \in \Xi_T]$, and $\Gamma_T = (K_t + \sigma^2 I)^{-1}$.

A kernel $k$ on triples $\xi = (x, z, y)$ can be defined by composing base kernels over instances $k_X$, explanations $k_Z$, and labels $k_Y$~\cite{krause2011contextual}.  Two standard combination operators are the \emph{direct sum} $\oplus$ and the \emph{tensor product} $\otimes$.  The direct sum of two kernels $k: \calX \to \bbR$ and $k': \calX' \to \bbR$ is a new kernel $(k \oplus k'): (\calX \times \calX') \to \bbR$:
\begin{align}
    (k \oplus k')((x_1, x'_1), (x_2, x'_2))
        & = k(x_1,x_2) + k(x'_1, x'_2)
        \label{eq:kernelsum}
\end{align}
The tensor product is defined analogously:
\begin{align}
    (k \otimes k')((x_1, x'_1), (x_2, x'_2))
        & = k(x_1,x_2) \cdot k(x'_1, x'_2)
    \label{eq:kernelprod}
\end{align}

GPs can provide a confidence on their own uncertainty. Hence, GPs are ideal for sampling regions in which the model is least certain about, with the purpose of increasing the confidence in that region. For this reason, GPs are well-suited for applications where querying the ground-truth reward function is expensive, like human-in-the-loop machine learning~\cite{guo2010gaussian} or black-box Bayesian optimization~\cite{thornton2013auto}.  \explearn exploits this property, too.

\begin{algorithm}[t]
    \caption{The \explearn algorithm}
    \label{alg:cgpucb}
    \begin{algorithmic}[1]
    \Require{Instances $\calX$, Explanations $\calZ$, Labels $\calY$, kernel over explanations $k$}
        \For{$t = 1, 2, \ldots, T$}
            \State Receive instance $x_t$ from user
            \State Select explanation $\hat{z}_t$ and prediction $\hat{y}_t$ using Eq.~\ref{eq:inference} \label{eq:play}
            \State Receive reward $f_t := f(x_t, \hat{z}_t, \hat{y}_t) + \epsilon_t$
        \EndFor
    \end{algorithmic}
\end{algorithm}

\explearn estimates the reward function by assuming a GP prior on $f$ and applies the \cgpucb contextual bandit algorithm to obtain an accurate estimate of the reward function interactively~\cite{krause2011contextual}.  The pseudo-code is listed in Algorithm~\ref{alg:cgpucb}.  \cgpucb selects arms $(\hat{z}_t, \hat{y}_t)$ to be presented to the user by balancing between \emph{exploitation} of arms with high mean reward and \emph{exploration} of arms with uncertain (high variance) reward in a principled manner.  In particular, an arm is chosen by maximizing
\[
    \textstyle
    \hat{z}_t, \hat{y}_t = \max_{z, y} \mu_t(x_t, z, y) + \sqrt{\beta_t} \sigma_t(x_t, z, y)
    \label{eq:inference}
\]
The first term prefers an arm with high mean reward, while the second  favors arms with high variance. The parameter $\beta_t$ strikes a balance between exploration and exploitation and should slowly increase over time to favor more exploration. With an appropriate choice of $\beta_t$, \cgpucb guarantees, under mild assumptions, no-regret as the number of iterations approach infinity: 

\begin{theorem}[\cite{krause2011contextual}]
    Let $\delta \in (0, 1)$, the norm $\|f\|_k$ be upper bounded by $\omega$, and the noise variables $\epsilon_t$ form a martingale difference sequence (that is, $\bbE[\epsilon_t \,|\, \epsilon_1, \ldots, \epsilon_{t-1}] = 0$ for all $t \ge 1$) uniformly bounded by $\sigma$.  Then, with probability at least $1 - \delta$, the cumulative regret of \cgpucb is upper bounded by $\calO^*(\sqrt{T \gamma_T \beta_T})$, where $\beta_t = 2\omega^2 + 300\gamma_t \ln^3(t / \delta)$ and $\gamma_t$ depends only on the choice of kernel.
    \label{thm:bound}
\end{theorem}

\noindent Here, $\calO^*$ denotes big-O notation with logarithmic factors suppressed and $\|f\|_k$ is the norm of the reward function $f$ w.r.t. the chosen kernel $k$.  In practice, convergence can often be improved by increasing $\beta_t$~\cite{krause2011contextual}.

Notice that the technical assumptions occur often in practice.  The martingale condition allows for correlated noise, which captures situations like annotation mistakes or temporary inattention.  On the other hand, as long as the kernel is expressive enough the reward function has finite norm and \cgpucb satisfies the no-regret property. Expressive kernels can be ensured through careful design, as discussed below.

\section{Explanations and Kernels}
\label{ssec:discussion}

\explearn is natively supports different kinds of explanations.  Here we discuss four representative forms of local explanations and their corresponding kernels.

\paragraph{Feature relevance}

A feature relevance explanation (FRE) $z$ is a $0$-$1$ vector whose $j$th element indicates whether the $j$th feature in $x_t$ is relevant or not.  For instance, a FRE for a diagnosis of pneumonia might highlight symptoms like ``fever'' and ``dry cough'' and exclude ones like ``skin rash''.  This kind of explanations is commonly associated with, e.g., decision trees and rule sets~\cite{freitas2014comprehensible,lakkaraju2016interpretable}.

In its simplest form, the space of possible FREs is the set of $0$-$1$ vectors with as many elements as the instances $x_t$. Higher-dimensional feature spaces are also supported.  Under the assumption that the user cares about how many (ir)relevant features were correctly identified by a FRE, the ground-truth reward will be proportional to the Hamming or Jaccard distance between a predicted FRE $\hat{z}$ and the ground truth $z$ for $x_t$.  A kernel that allows to encode this kind of reward is the set kernels.

\paragraph{Feature importance}  An explanation $z$ is a real-valued vector in which the $j$th element represents the degree of agreement of $j$th towards the target decision.  In the simplest case, elements are simply $0$, $1$, and $-1$.  In our example, ``fever'' and ``dry coughing'' would be associated to a positive score and ``skin rash'' to a negative one.  This kind of explanation is commonly associated to (discretized, sparse) linear models and neural nets (e.g., saliency maps).

In feature importance the user provides a score for the relevance of each feature. A ground-truth model outputs real valued explanations scaled for convenience in the $[0,1]$ interval, where $1$ indicates the maximum relevance. In this case the user cares of the overall score of each feature. As such, the ground-truth reward of the explanation is the cosine similarity $z^\top\hat{z}$ between the real and the predicted explanation.
The choice of the kernel is more flexible, and depends on the data and the domain. Common choices for kernels are real-valued kernels, such as linear, polynomial, and RBF kernels.

\paragraph{Feature ranking} An explanation $z$ is a (sparse) preference relation that specifies the relative importance of different features. In this case, the importance value of each feature individually does not matter as long as the importance order is unaltered. This form of explanation is more interpretable for users as it eschews from checking the absolute value of each feature.  In our example, ``fever'' could be ranked higher than ``dyspnea''.  If the opposite was true, a different condition would be diagnosed, for instance common flu.

In feature ranking, the user provides a preference order for features.
A ground-truth model outputs a natural number that corresponds to the position of each feature in the explanation vector. As the user is concerned only with the relative position of the features, the reward can be described as the amount of pairwise agreement among the ground-truth ranking $z$ and the model's explanation ranking $\hat{z}$. The amount of agreement can be expressed as the (inverse) Kendall distance: 
$$
    1 - \sum_{1 \leq i, j \leq n} \mathbbm{1}(\pi(z(i)) < \pi(\hat{z}(j)) \land \pi'(z(i)) < \pi'(\hat{z}(j)))
$$
where $\pi(z)$ maps feature values into positions and $\mathbbm{1}$ is the indicator function. The Kendall distance is usually normalized between in the $[0,1]$ interval, by dividing with the number of pairs $n(n{-}1)/2$. 
Similarly, the (inverse) Kendall distance is a kernel for explanations~\cite{jiao2015kendall}.

\paragraph{Feature traces} An explanation $z$ is the sequence of steps that leads to the classifier prediction for an instance $x$. In this case, the overall sequence is important in explaining the behaviour of a classifier. This kind of explanations are suitable for order-dependant predictor such as decision trees (DTs) or logic programs. A DT, for instance, justifies a diagnosis via a decision path from the root to the leaf. In the medical case, a DT could verify first that temperature is below $37$, and if false verify whether it is above $39$ and only then output ``high-risk pneumonia''. Traces allow a user to verify the algorithm steps in the prediction in a fine-grained manner.

In feature traces, the user provides a sequence of steps. A ground-truth model outputs a ordered subset of features that corresponds to a specific trace. The user in this case checks the sequence starting from the root and the model receives a reward proportional to the longest sub-sequence matching the ground-truth sequence. Suitable kernels for feature traces is the family of kernels on proof trees~\cite{passerini2006kernels}.

\subsection{Discussion}

Is learning to explain a sound learning task?  If so, is \explearn suited for it?  In the following, we argue that both questions find an affirmative under reasonable assumptions.

Before proceeding, we start by assuming that the user's rewards are noisy but drawn from a fixed, ground-truth reward function $f: \calX \times \calZ \times \calY \to \bbR$.  This assumption is quite general and underlies most human-in-the-loop applications of GPs~\cite{guo2010gaussian,wilson2015human}.\footnote{In some cases (like concept or preference drift) users can be inconsistent and thus violate this assumption.  This issue, however, is orthogonal to our contribution.}

Theorem~\ref{thm:bound} states that \cgpucb can achieve no-regret as long as a kernel $k$ exists such that $f$ has low norm in the corresponding reproducing kernel Hilbert space (RKHS).  The question becomes whether such a kernel exists.  One way to ensure that $f$ has low norm w.r.t. $k$ is to consider a \emph{universal} kernel~\cite{steinwart2001influence}.  Such kernels, which include the RBF kernel, guarantee that \emph{all} functions, including $f$, can be approximated arbitrarily well in the corresponding RKHS\footnote{Universality is usually defined for continuous data.  A discrete counterpart is discussed in~\cite{sriperumbudur2011universality}.}.  This entails, on the one hand, that GPs equipped with a universal kernel can approximate any reward function $f$, and on the other, that \explearn will eventually acquire a high-quality estimate of $f$.

Designing a universal kernel $k$ over triples $(x, z, y)$ is not trivial, especially if additional modeling assumptions have to be taken into account~\cite{gretton2015simpler}.  A more straightforward alternative is to define universal kernels $k_X$, $k_Z$, and $k_Y$ over instances, explanations, and labels and then compose them using a combination of tensor products and/or direct sums so that the resulting kernel $k_{XZY}$ is also universal.  It turns out that direct sum, however, is not viable in this sense.  Indeed, any reward function $f_\oplus: \calX \times \calX' \to \bbR$ defined in the RKHS of $k \oplus k'$ can be written as:
\begin{align*}
    f_\oplus(x, x')
        & \textstyle = \sum_{i=1}^n \alpha_i \,\cdot\, (k \oplus k')((x,x'), (x_i, x'_i))
    \\
        & \textstyle = \sum_{i=1}^n \alpha_i k(x, x_i) + \sum_{i=1}^n \alpha_i k'(x', x'_i))
    \\
        & \textstyle = f(x) + f'(x')
\end{align*}
where $n \in \bbN$, $\alpha_1, \ldots, \alpha_n \in \bbR$, the $x_i$'s belong to $\calX$, the $x'_i$'s to $\calX'$, and $f$ and $f'$ are defined in the RKHS of the base kernels $k$ and $k'$, respectively.  In other words, the direct sum imposes \emph{additive independence} between the elements of $\calX$ and the elements of $\calX'$.  Consider a kernel over triples $(x, z, y)$ obtained using a direct sum, e.g., $k_{XZY} = k_X \oplus k_Z \oplus k_Y$.  In this case, additive independence entails that the ranking of alternative explanations is independent of the instance and label.  Indeed, any reward function $f_{XZY}(x,z,y)$ in the RKHS of $k_{XZY}$ satisfies:
\begin{align*}
    \forall x, y, z, z' \,.\, f_Z(z)
        & \ge f_Z(z') \iff
    \\
    f_{XZY}(x,z,y)
        & \ge f_{XZY}(x,z',y)
\end{align*}
As a consequence, an explanation that is optimal for a given instance and label is still optimal for completely different instances and labels.  This is not always desirable.

It follows that the direct sum does not preserve universality: even if $k$ and $k'$ are universal kernels, $k \oplus k'$ is not, as it cannot express reward functions that are not additively independent.  In stark contrast, the tensor product does preserve universality:  if $k$ and $k'$ are universal, so is $k \otimes k'$~\cite{blanchard2011generalizing,gretton2015simpler,szabo2017characteristic}.  Hence, it is straightforward to build universal kernels $k = k_X \otimes k_Z \otimes k_Y$ that ensure with high probability that \explearn attains no-regret.  In practice, simpler, non-universal kernels can enjoy better sample complexity.  A good trade-off between expressiveness and convergence could be achieved by leveraging automated kernel search, cf.~\cite{grosse2012exploiting} and follow-ups.

\subsection{Advantages and Limitations}

\explearn has a number of important features.  First, it is the first algorithm that sports learning guarantees for learning from explanations.  Second, the models it learns are self-explainable and---unlike post-hoc explainers based on model distillation, like LIME~\cite{ribeiro2016should}---they output exact explanations.

Many aspects of \explearn can be improved.  For instance, \explearn relies on scoring feedback, which might not be very natural for end-users.  This limitation can be lifted by, e.g., asking for ranking feedback~\cite{guo2010gaussian}.  More importantly, the space of explanations is combinatorial in nature.  It follows that inferring a high-reward arm can be computationally non-trivial:  depending on the kind of explanations and kernels used, Eq.~\ref{eq:inference} can become an {\NPhard} combinatorial optimization problem.  Efficient solvers do exist for special cases, e.g., when the kernel is linear.  A more general, yet efficient, alternative is to solve the arm selection problem by adapting recent techniques for combinatorial bandits~\cite{koriche2018compiling,sakaue2018efficient}.  This is, however, non-trivial for general kernels.  A simpler option is to solve Eq.~\ref{eq:inference} only approximately, leading to practical algorithms for more complex explanation types.

\section{Experiments}
\label{sec:experiments}

We present preliminary evidence that \explearn delivers on its promises.  In particular, we evaluate whether \explearn can \textbf{(Q1)} learn from feedback on both explanations and labels, \textbf{(Q2)} learn from and output different kinds of explanations, \textbf{(Q3)} outperform a random sampling strategy.  To this end, we implemented \explearn using Python 3 and scikit-learn and ran experiments on a 24 cores Intel(R) Xeon(R) CPU E5-2440 2.40GHz, 192GiB RAM.  The experiments can be found at \url{https://git.io/JL0Sv}.

\spara{Kernels.}  For simplicity, we use RBF kernels for instances, explanations, and labels, and combine them either into a \textsc{prod} kernel $k = k_X \otimes k_Z \otimes k_Y$ or a \textsc{sum} kernel $k = (k_X \oplus k_Z) \otimes k_Y$, without any fine-tuning.  As discussed above, fine-tuning  could in principle attain better results.

\spara{Baseline and metrics.} Besides the \explearn algorithm, which uses the \cgpucb arm selection strategy (as per Eq.~\ref{eq:inference}), we implemented a ``Random'' baseline that asks for feedback on random arms.  Performance is measured in terms of average cumulative regret, see Eq.~\ref{eq:cumreg}.  To ensure a fair comparison, the cumulative regret is computed relative to the best arm (that is, $\argmax_{z,y} f(x_t, z, y)$) rather than the arm shown to the user $(\hat{z}_t, \hat{y}_t)$.  Besides the random baseline, we test the \textsc{prod} and \textsc{sum} kernel. Although UCB generally performs better than Random, we believe our setup requires a more thorough investigation.

\subsection{Datasets and Explanations} 

We evaluate \explearn on two datasets with known ground-truth explanations and reward function. The ground-truth explanations are either synthesized or extracted from a pre-trained interpretable model.

\spara{Colors}~\cite{ross2017right} dataset consists of $5{\times}5$ images labeled as positive or negative according to one of two possible rules:  rule 1 holds if the four corner pixels have the same color, while rule 2 holds if the three top-middle pixels have different colors.  All images satisfy both rules or neither of them, making it impossible to disambiguate between the two rules based on labels alone.  Both feature relevance and feature importance \textbf{ground-truth explanations} were generated, as follows.  Feature relevance explanations are $0$--$1$ vectors that identify those pixels relevant to the ground-truth label, namely the four corner pixels for rule 1 and the three top middle pixels for rule 2.  Feature importance explanations are similar, except that the relevant pixels are weighted as $+1$ or $-1$ depending on whether they support or conflict with the label.  Notice that feature importance explanations depend on the both image and the label.  \explearn was tested on both kinds of explanations.  The models outputs feature relevance or feature importance explanations depending on the supervision.  The \textbf{reward} is taken to be the cosine similarity between the ground-truth and the predicted explanation multiplied by $1$ if the predicted and real label agree and by $-1$ otherwise.

\spara{Banknote}~\cite{dua2019uci} consists of 4-dimensional instances extracted from images of counterfeit and genuine banknotes.  The data set includes 1372 instances, 610 counterfeit and 762 genuine. The \textbf{ground-truth explanation} for each instance is the path of a decision tree with depth between 5 and 10. The corresponding \textbf{model explanation} is any set of feature conditions in a root-to-leaf path.  The \textbf{reward} is the Jaccard similarity among the ground-truth and the model explanation multiplied multiplied by $1$ if the predicted and real label agree and $-1$ otherwise.

\subsection{Results and Discussion}

\begin{figure}[tb]
\centering
    \begin{subfigure}{.23\textwidth}
      \centering
      \includegraphics[width=\textwidth]{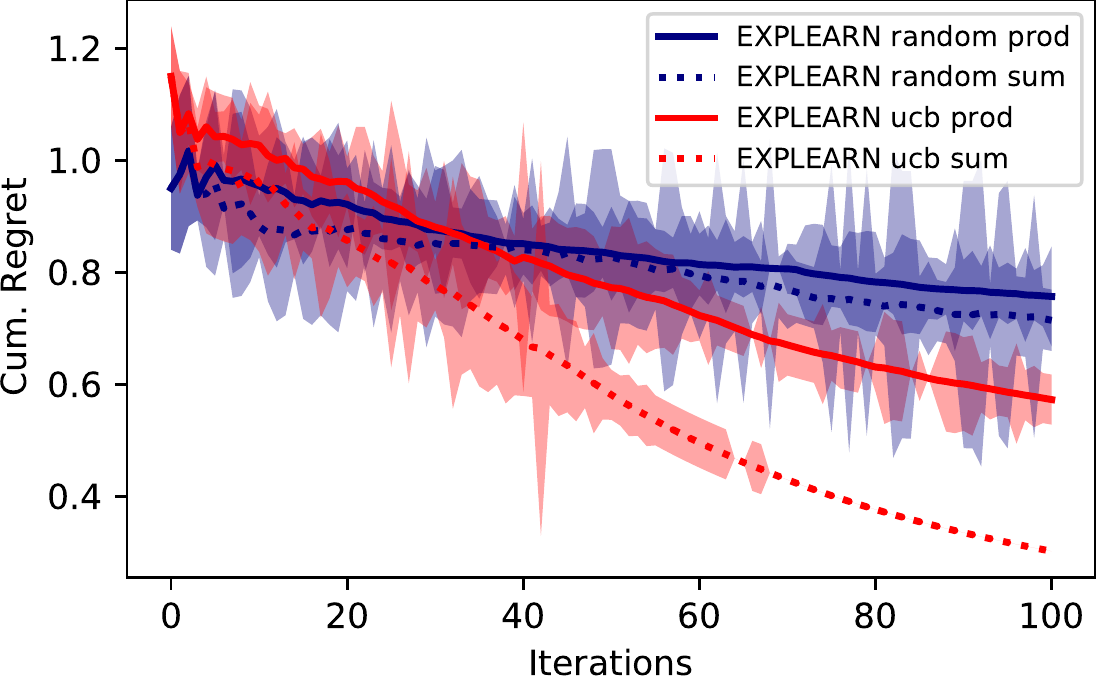}
      \caption{Feature relevance}
      \label{fig:colors-0-rel}
    \end{subfigure}
    \begin{subfigure}{.23\textwidth}
      \centering
      \includegraphics[width=\textwidth]{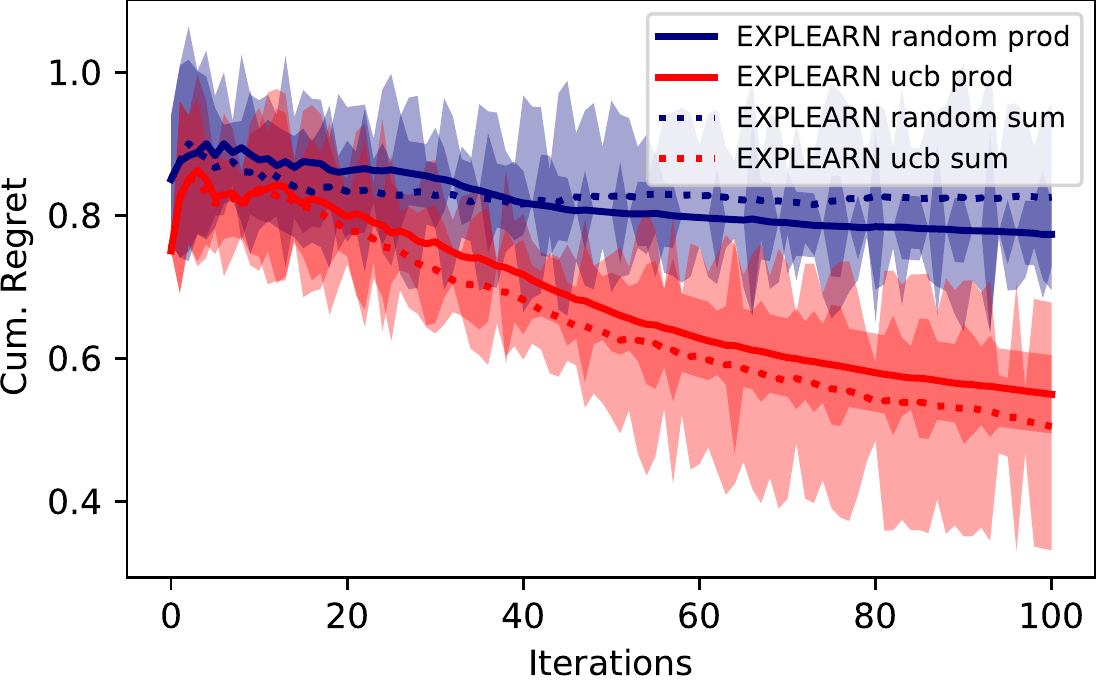}
      \caption{Feature importance}
      \label{fig:colors-0-pol}
    \end{subfigure}
    \caption{Cumulative regret on \colors dataset on \textbf{rule 0}.}
    \label{fig:colors-0}
\end{figure}

\begin{figure}[tb]
\centering
    \begin{subfigure}{.23\textwidth}
      \centering
      \includegraphics[width=\textwidth]{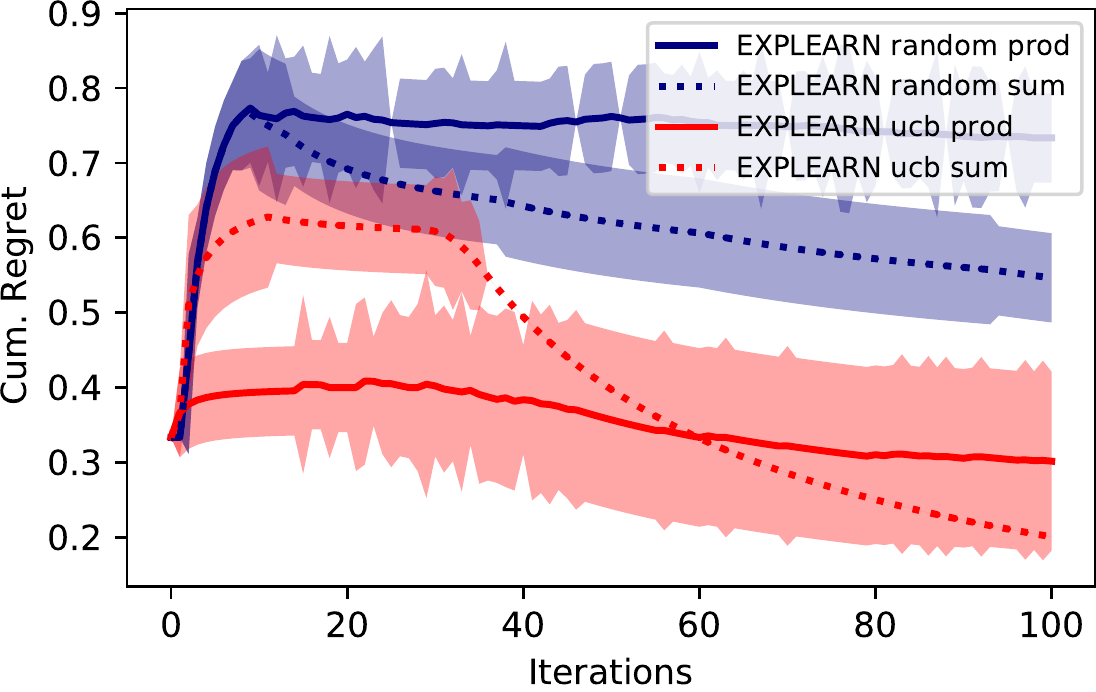}
      \caption{Feature relevance}
      \label{fig:colors-1-rel}
    \end{subfigure}
    \begin{subfigure}{.23\textwidth}
      \centering
      \includegraphics[width=\textwidth]{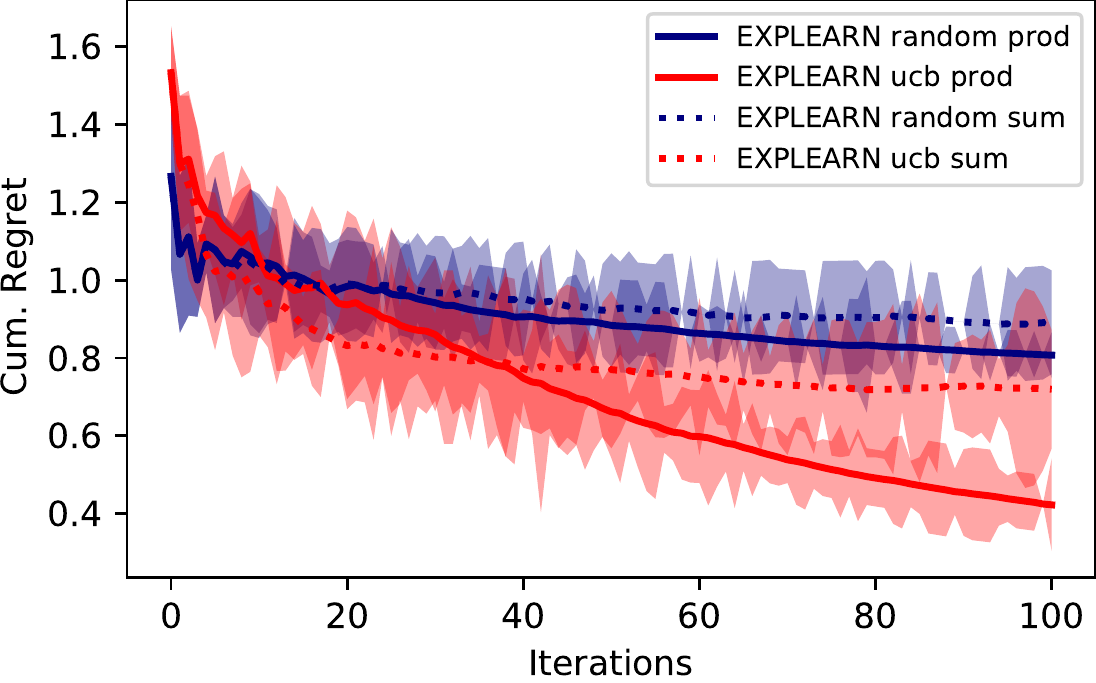}
      \caption{Feature importance}
      \label{fig:colors-1-pol}
    \end{subfigure}
    \caption{Cumulative regret on \colors dataset on \textbf{rule 1}.}
    \label{fig:colors-1}
\end{figure}

\begin{figure}[tb]
  \centering
  \includegraphics[width=.6\columnwidth]{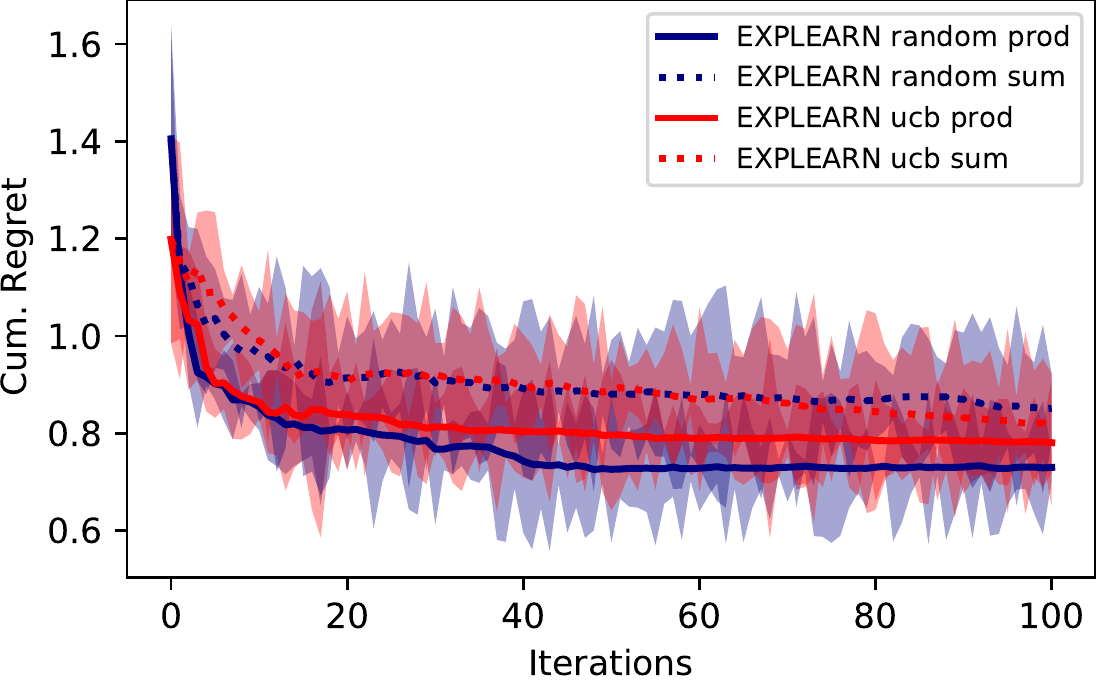}
  \caption{\banknote dataset, trace reward.}
  \label{fig:banknote}
\end{figure}

Figures~\ref{fig:colors-0}--\ref{fig:banknote} report the average and the standard deviation (in shaded color) of the cumulative regret as a function of the number of iterations. Recall that in each iteration the algorithm receives an instance, selects an explanation and a prediction and receives a reward. We report the results for the random sampling strategy and the UCB sampling strategy. We also report results for both the the sum and product kernel combination.

\noindent\textbf{Q1: Does \explearn learn from explanations?}
The first and foremost question is whether \explearn can learn from feedback on explanations and labels. The plots show that the regret consistently decreases with the number of iterations / examples in all datasets, as expected. However, the results vary among different explanation types and datasets. In particular, for banknote (Figure~\ref{fig:banknote}) the regret falls fast in the first iterations but reaches a plateau around 60 iterations. The reason is probably the way explanations are constructed which disregard the order.
%
\noindent\textbf{Q2: Does \explearn work with different kinds of explanations?} 
To test the ability of \explearn to learn from different explanation models we experiment with feature relevance (Figure~\ref{fig:colors-0-rel}--\ref{fig:colors-1-rel}), feature importance (Figure~\ref{fig:colors-0-pol}--\ref{fig:colors-1-pol}), and feature traces~\ref{fig:banknote}). In all cases \explearn learns from explanations, although results are different. On the \colors dataset, with sparse features and explanations the sum kernel performs generally better, as the pixel in those images are not correlated. As such, the independence assumption is more realistic. On the \banknote dataset, all the variants perform similar, although the product kernel outperform the sum kernel. In the case of decision trees features are correlated. Therefore, the independence assumption of sum fails to capture such correlations. 
\noindent\textbf{Q3: Is UCB better than random?}
Finally, we look at the question whether UCB is a better strategy than random. In general, the experiments confirm the theoretical advantage of UCB. On the other hand, a random exploration of new points might be beneficial in particularly challenging datasets a more complicated explanations, as in the case of \banknote in Figure~\ref{fig:banknote}. However, such a behaviour might be determined by the lack of ordering in the reward model we applied in \banknote.

\section{Related Work}\label{sec:related-work}

Roughly speaking, explainable machine learning focuses on two major problems:  designing white-box models that are inherently interpretable (cf.~\cite{lakkaraju2016interpretable}) and understanding black-box models and their predictions~\cite{craven1996extracting,guidotti2018local}.  These forms of transparency are a prerequisite to identifying bugs and biases in learned models, but they are also insufficient to \emph{correct} those models.  In contrast, \explearn explicitly imposes supervision on the explanations themselves.

Our work is strongly related to explanatory interactive learning~\cite{teso2019explanatory,schramowski2020right,lertvittayakumjorn2020find} which integrates explanations into pool-based active learning~\cite{settles2009active} (with some exceptions~\cite{popordanoska2020machine}).  In XIL, the machine picks instances and presents prediction and explanations for them to the user, who then supplies feedback on both the label and the explanation.  This enables the machine to learn models that are ``right for the right reasons''~\cite{ross2017right}.  \explearn also relies on explanatory interaction, but it tackles sequential learning tasks in which instances arrive over time and are not chosen by the machine.  In contrast to XIL, \explearn also sports sound learning guarantees.

Most closely related to our work, self-explainable neural networks (SENNs)~\cite{alvarez2018towards} natively output gradient-based explanations for their decisions and have been combined with explanatory active learning~\cite{teso2019toward}.  The models learned by \explearn are also self-explainable, but they are not restricted to gradient-based explanations.  In addition, SENNs assume that explanations vary smoothly (in a Lipschitz sense), which is incorrect near the decision boundary.  In contrast, \explearn allows to more freely control how explanations generalize by choosing an appropriate kernel function.

\section{Conclusion}\label{sec:conclusion}

We introduced \explearn, an approach for learning to explain by interacting with a human supervisor.  \explearn acquires self-explainable predictors that output explanations compatible with those supplied by their human supervisor.  This is achieved by combining Gaussian Processes with a principled contextual bandit algorithm.  Preliminary experiments on real-world and synthetic data sets indicate that, given enough examples, \explearn successfully acquires models that output high-reward predictions and explanations.  Of course, a more thorough evaluation of \explearn is due.  Another high-priority task is to speed up inference of the best arm by leveraging, e.g., knowledge compilation techniques used in combinatorial bandits~\cite{koriche2018compiling,sakaue2018efficient}.  Other research directions involve strengthening the semantic consistency of instances-explanation-prediction triples by introducing specialized kernels, and enabling run-time customization of the kind and complexity of the explanations output by the model suitably to the user's needs, as done in~\cite{lage2018human}.

\clearpage 
\bibliography{paper}
\end{document}